\documentclass{article}

\usepackage{PRIMEarxiv}

\usepackage[utf8]{inputenc} 
\usepackage[T1]{fontenc}    
\usepackage{hyperref}       
\usepackage{url}            
\usepackage{booktabs}       
\usepackage{amsfonts}       
\usepackage{nicefrac}       
\usepackage{microtype}      
\usepackage{lipsum}
\usepackage{fancyhdr}       
\usepackage{graphicx}       
\graphicspath{{media/}}     

\title{Exploring Protein Language Model Architecture-Induced Biases for Antibody Comprehension}

\author{
  Mengren (Bill) Liu\\
  Department of Materials Science and Engineering \\
  Massachusetts Institute of Technology \\
  MA 02139, USA\\
  \texttt{billliu@mit.edu} \\
   \And
  Yixiang Zhang \\
  College of Electrical Engineering\\
  Zhejiang University \\
  Zhejiang 310058, China\\
  \texttt{12110023@zju.edu.cn} \\
   \And
  Yiming (Jason) Zhang$^{\ast}$ \\
  Department of Biological Engineering \\
  Massachusetts Institute of Technology \\
  MA 02139, USA\\
  \texttt{ymjzhang@mit.edu} \\
}

\begin{document}
\maketitle

\begin{abstract}
Recent advances in protein language models (PLMs) have demonstrated remarkable capabilities in understanding protein sequences. However, the extent to which different model architectures capture antibody-specific biological properties remains unexplored. In this work, we systematically investigate how architectural choices in PLMs influence their ability to comprehend antibody sequence characteristics and functions. We evaluate three state-of-the-art PLMs—AntiBERTa, BioBERT, and ESM2—against a general-purpose language model (GPT-2) baseline on antibody target specificity prediction tasks. Our results demonstrate that while all PLMs achieve high classification accuracy, they exhibit distinct biases in capturing biological features such as V gene usage, somatic hypermutation patterns, and isotype information. Through attention attribution analysis, we show that antibody-specific models like AntiBERTa naturally learn to focus on complementarity-determining regions (CDRs), while general protein models benefit significantly from explicit CDR-focused training strategies. These findings provide insights into the relationship between model architecture and biological feature extraction, offering valuable guidance for future PLM development in computational antibody design.
\end{abstract}

\keywords{Protein Language Models \and Representation Learning}

\section{Introduction}

\label{intro}

Proteins, comprising sequences of 20 distinct amino acids (AAs), constitute the fundamental functional units of biological systems. The sequential arrangement of these AAs generates diverse protein structures that facilitate cellular functions. This inherent complexity in protein sequences has led to the emergence of computational approaches that treat protein sequences as a form of biological language, amenable to analysis through advanced language modeling techniques.

\begin{figure}[ht]
\vskip 0.2in
\begin{center}
\centerline{\includegraphics[width=\columnwidth]{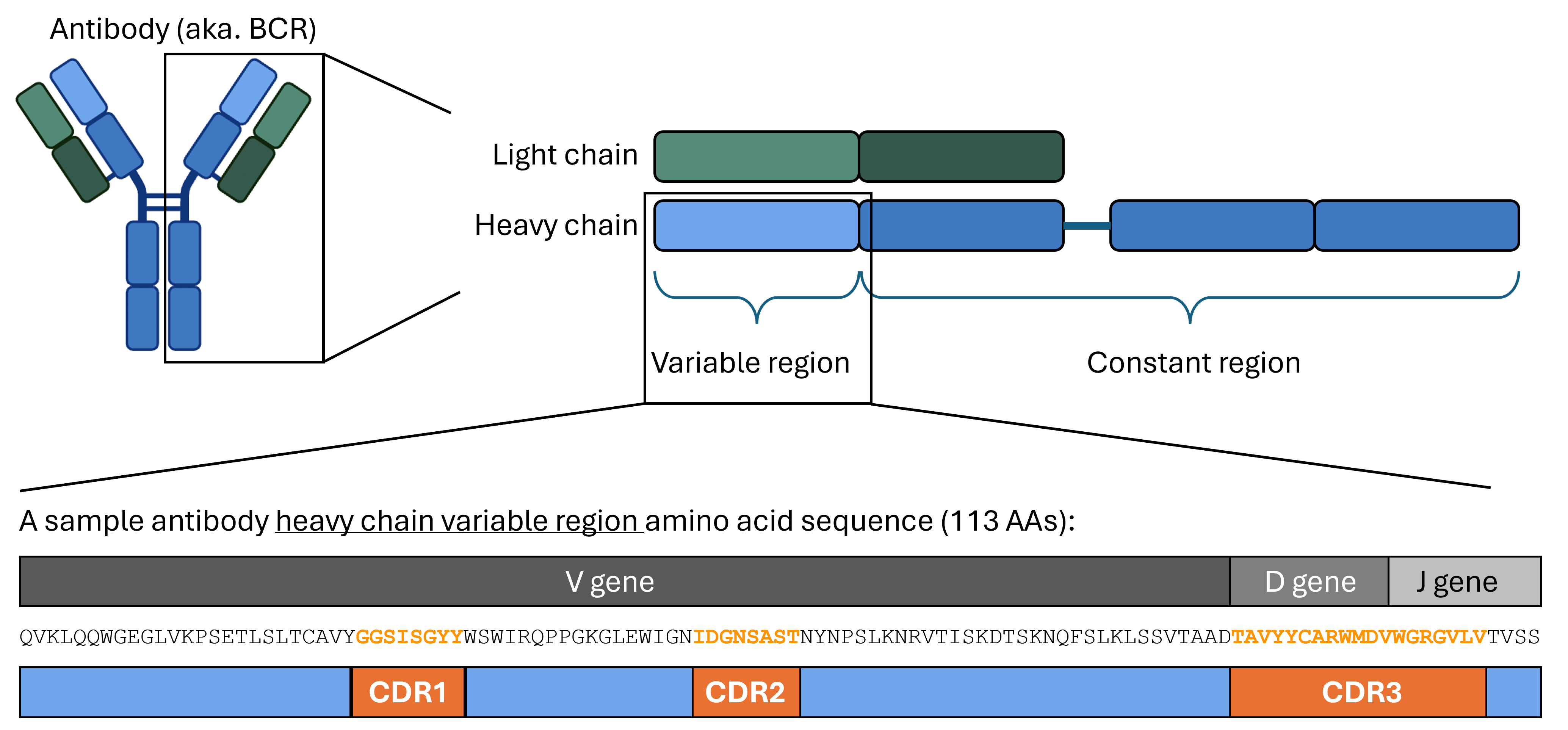}}
\caption{A schematic illustration of the composition of an antibody. Antibody is composed of two heavy chains and two light chains. Both chains have a variable region that is unique to each B cell and a constant region (one of IgM, IgD, IgG, IgA, and IgE for heavy chains, and either IgK or IgL for light chains). The variable region is generated by the genetic combination of a V gene, a D gene, and a J gene. The complementary-determining regions (CDRs) on the variable region are the most important segments that determine antigen specificity.}
\label{BCR}
\end{center}
\vskip -0.2in
\end{figure}

Recent advances in natural language processing (NLP) have enabled parallel developments in protein sequence analysis. While NLP models excel at comprehending human language patterns, protein language models (PLMs) have been specifically developed to encode and interpret the biological information embedded within protein sequences. Among the various protein classes available for investigation, antibodies present a particularly compelling case study due to their complex sequence-structure-function relationships.
Antibodies, serving as crucial components of the adaptive immune system, exhibit remarkable specificity in antigen recognition. Prior to secretion, these molecules exist as B cell receptors (BCRs) on B cell surfaces. The exceptional diversity of antibody sequences arises from sophisticated molecular mechanisms, primarily VDJ recombination and somatic hypermutation (Figure \ref{BCR}).

VDJ recombination represents a programmed genetic rearrangement process that generates antibody diversity through the combinatorial assembly of variable (V), diversity (D), and joining (J) gene segments. This process creates a variable region that determines antigen specificity, while maintaining a constant region that defines effector functions. Following antigen encounter, B cells undergo somatic hypermutation, introducing point mutations within the variable region to optimize antigen binding affinity.

The complementarity-determining regions (CDR loops) constitute the most critical elements of antibody structure. These loops, comprising three regions each in the heavy and light chains, form the antigen-binding interface and determine specificity (Figure \ref{BCR}). The therapeutic potential of antibodies has been well-established \cite{Leavy2010TherapeuticFuture,Paul2024CancerAntibodies}, though the design and discovery process remains resource-intensive \cite{Chiu2016EngineeringTherapeutics,Lu2020DevelopmentDiseases}.

Transformer-based architectures currently represent the state-of-the-art in protein language modeling. BERT-based models, including BioBERT \cite{lee2020biobert}, have been adapted for biological sequence analysis. AntiBERTa \cite{barton2023antiberta2} represents a significant advancement as the most comprehensive antibody-specific language model to date. Evolutionary Scale Modeling (ESM) approaches, exemplified by ESM2 \cite{lin2022evolutionary}, leverage extensive protein datasets to capture evolutionary relationships, offering insights into protein structure and function that could advance antibody engineering efforts.

This study aims to systematically evaluate how different PLM architectures, employing distinct representation learning strategies, capture antibody-specific properties and characteristics based solely on amino acid sequences. Our investigation focuses on understanding architecture-induced biases and their implications for antibody sequence comprehension.

\section{Method}
\label{method}
\begin{figure}[ht]
\vskip 0.2in
\begin{center}
\centerline{\includegraphics[width=\columnwidth]{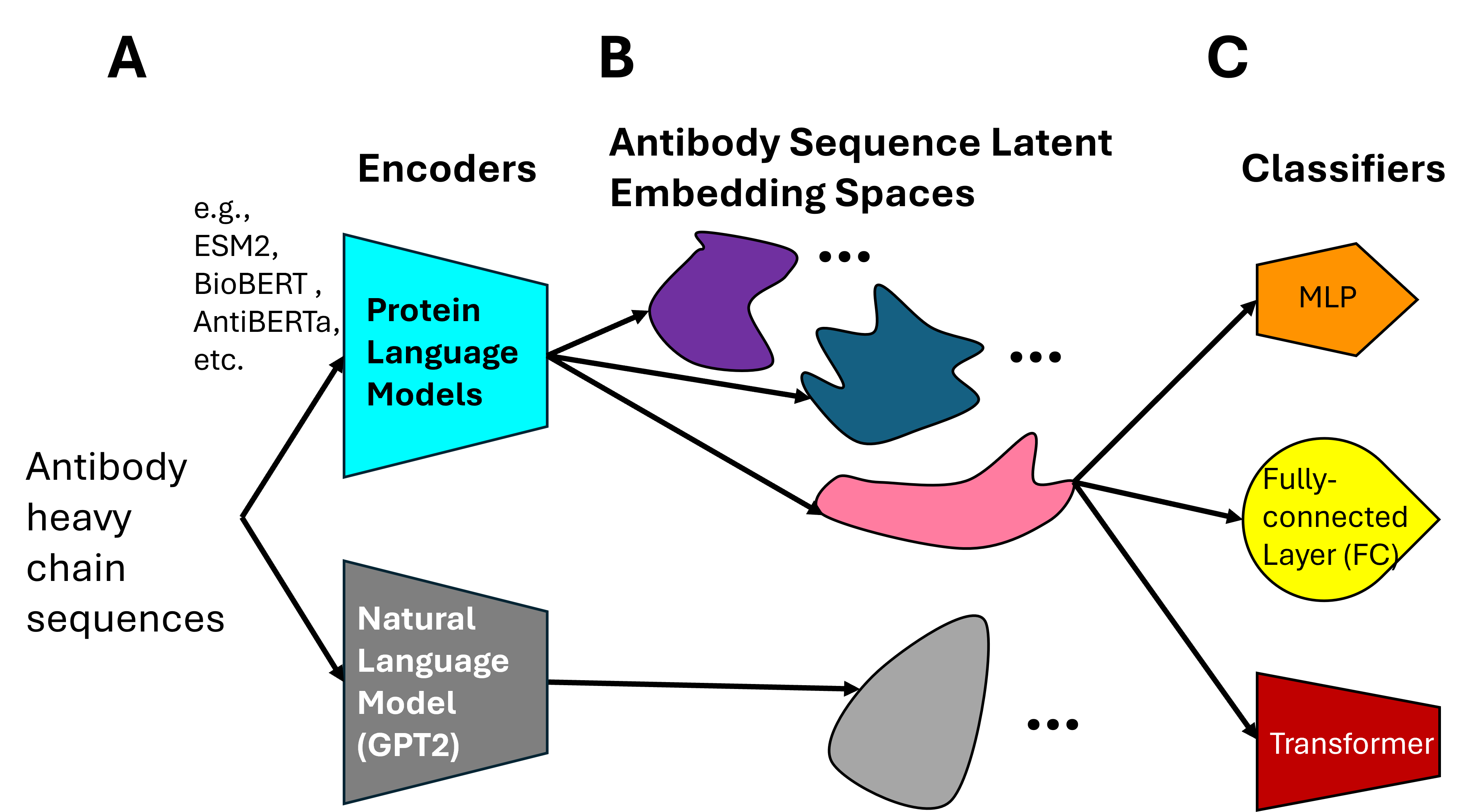}}
\caption{Graphical abstract of model architectural exploration. \textbf{A.} Encoding antibody sequences into latent embeddings using different encoders (i.g., ESM2, AntiBERTa, etc.). \textbf{B.} Adding antibody target labels to each embedded antibody sequence. \textbf{C.} Train classifiers of different architectures for each embedding to predict antigen binding.}
\label{overview}
\end{center}
\vskip -0.2in
\end{figure}

Our experimental framework employs a dataset comprising Rhesus macaque antibody heavy chain variable region sequences, stratified into three categories: 3,459 HIV-specific sequences, 3,459 Pneumococcus serotype 3 (Pn3)-specific sequences, and 3,459 non-Pn3-specific sequences. The classification pipeline consists of three primary components (Figure \ref{overview}):

1. Embedding Generation: Antibody sequences were processed through PLMs to generate sequence embeddings. To serve as a baseline, a general-purpose pre-trained text tokenizer (GPT-2 \cite{radford2019language}) was used as a control, as it is unlikely to capture the biological functions of antibodies. The parameters of the language models are shown in Table \ref{para}.

\begin{table}[t]
\caption{Language model parameters.}
\label{para}
\vskip 0.15in
\begin{center}
\begin{small}
\begin{sc}
\begin{tabular}{p{2cm}p{3cm}p{3cm}p{3cm}p{3.5cm}}
\toprule
Model & Number of Layers & Hidden State Size & Number of Attention Heads & Architecture\\
\midrule
BioBERT & 12 & 768 & 12 & BertModel\\
\addlinespace[0.5em]
AntiBERTa & 16 & 1024 & 16 & RoFormerForMaskedLM\\
\addlinespace[0.5em]
ESM2 & 12 & 480 & 20 & EsmForMaskedLM\\
\addlinespace[0.5em]
GPT-2 & 12 & 768 & 12 & GPT2LMHeadModel\\
\bottomrule
\end{tabular}
\end{sc}
\end{small}
\end{center}
\vskip -0.1in
\end{table}

2. Metadata Pairing: Each antibody sequence embedding was paired with its corresponding binding target information.

3. Classification: A classifier network was constructed to predict binding specificity (anti-HIV or anti-Pn3) taking the embeddings as the input. Various classifier architectures were explored, including a Transformer, a fully connected (FC) layer, and a 3-layer multilayer perceptrons (MLP). For the fully connected and MLP classifiers, sequence embeddings were pooled using the last hidden state of the PLMs. For the transformer-based classifier, sequence embeddings were pooled at the output of the transformer classifier, just before the activation layer.

Model optimization involved fine-tuning the classifier layers and the terminal two layers of the PLMs to enhance task-specific feature extraction. All experiments were conducted using Google Colab's T4 GPU infrastructure. Training utilized a batch size of 128 sequences over approximately 30 epochs.

\section{Results and Discussion}
\label{rd}

\subsection{Can PLMs be used for antibody target specificity prediction?}
\begin{figure}[ht]
\vskip 0.2in
\begin{center}
\centerline{\includegraphics[width=\columnwidth]{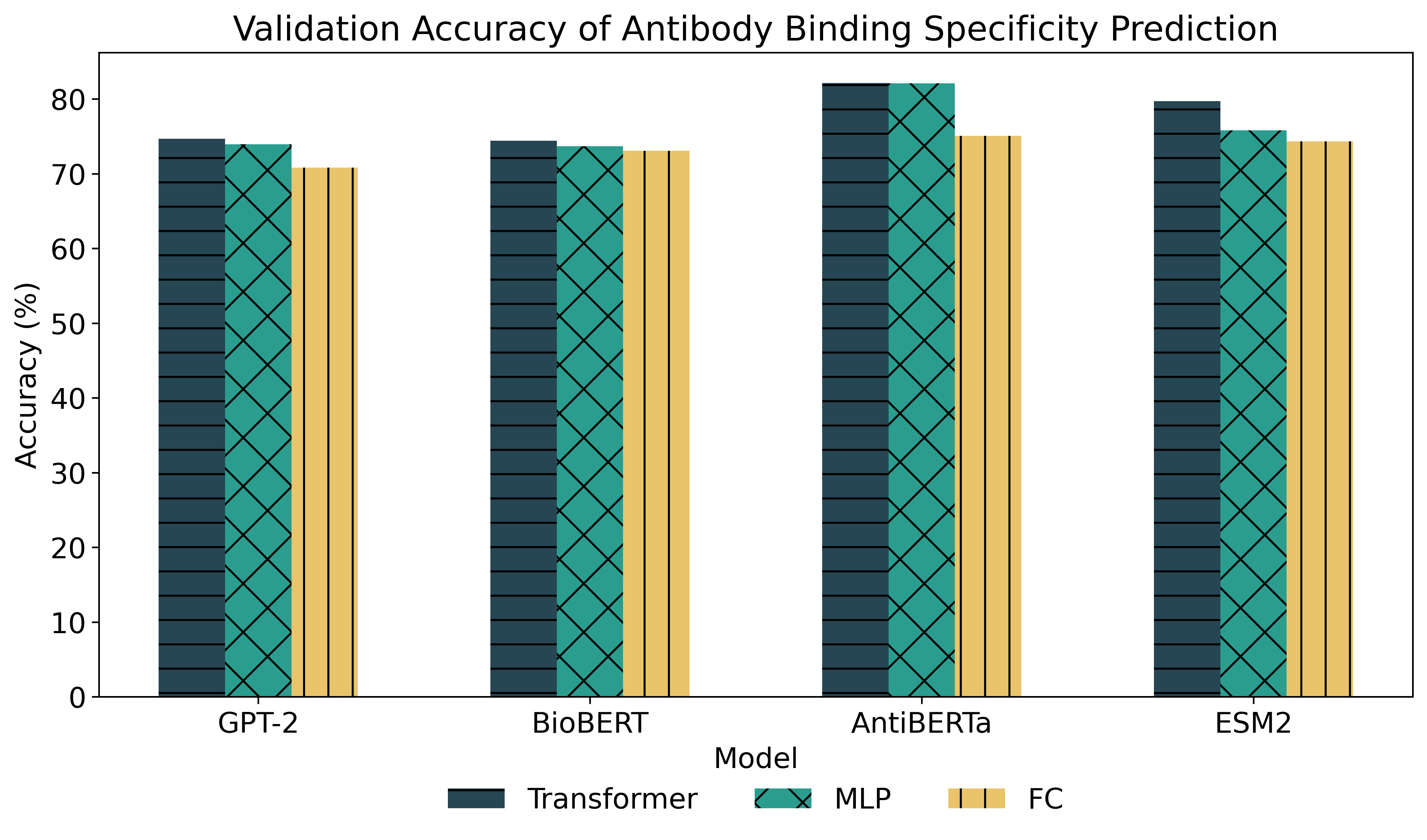}}
\caption{The validation accuracy percentages of each classifier architecture paired with corresponding PLMs or GPT-2 models. Grouped bars differentiate by harch patterns and dark green (the transformer classifier), turquoise (the MLP classifier), and yellow (the FC classifier).}
\label{compare}
\end{center}
\vskip -0.2in
\end{figure}

Based on our experimentation, the answer is YES! As shown in Figure \ref{compare}, all PLMs excelled at classifying antibody target specificity based on AA sequences regardless of the classifier architecture. Even the general-purposed NLP model GPT-2 resulted in a decent accuracy at classifying antibody targets. With similar and limited computation time, the transformer classifier consistently achieved the highest accuracy with varying margins. The MLP classifier came second, and the FC classifier demonstrated the lowest accuracy among the three. This suggests that transformer-based architectures are powerful tools for executing classification tasks, potentially thanks to their ability to capture the long-range context of the three complementary determining regions (CDRs) of an antibody sequence.

\begin{figure}[ht]
\vskip 0.2in
\begin{center}
\centerline{\includegraphics[width=\columnwidth]{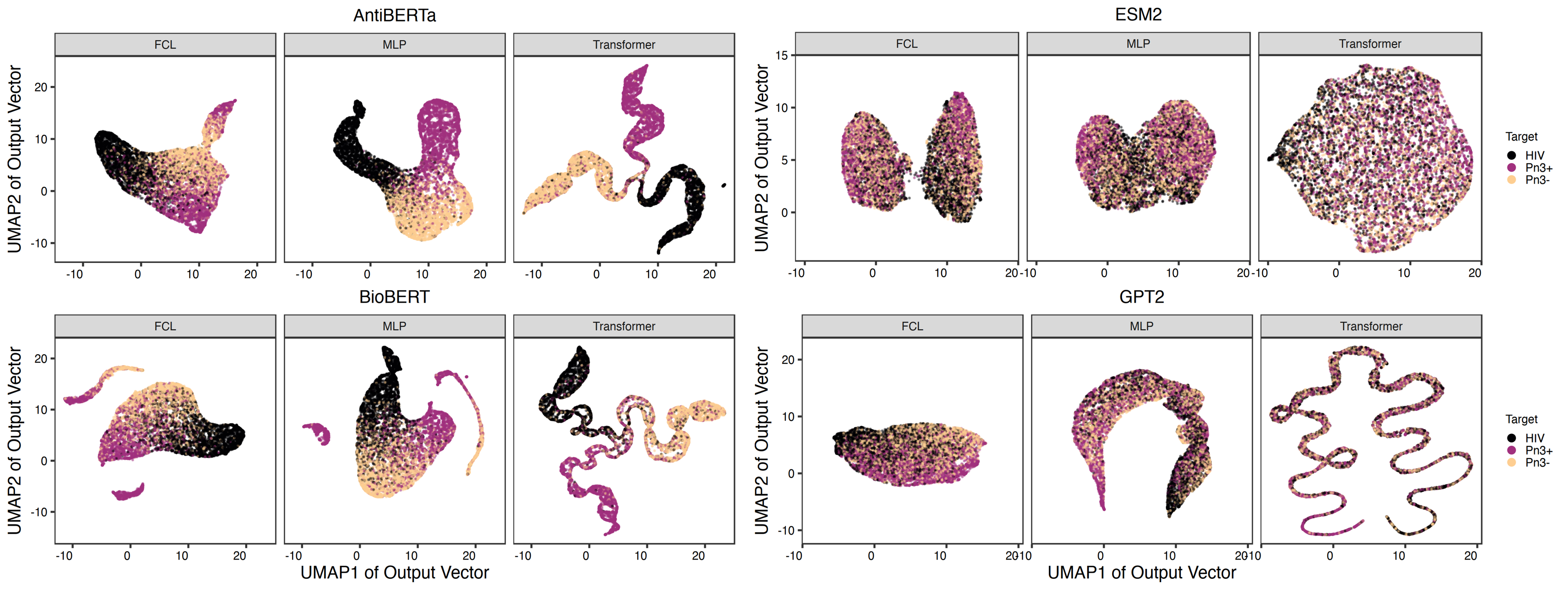}}
\caption{UMAP of output vectors under different language models and classifier model pairs. The output vector of each antibody sequence was colored based on their target specificity in black (HIV+), purple (Pn3+), and beige (Pn3-).}
\label{UMAP_output}
\end{center}
\vskip -0.2in
\end{figure}

Seeing is believing right? The Uniform Manifold Approximation and Projection (UMAP) is a powerful dimensionality reduction tool, and we used it to directly visualize the outputs of our models by projecting them onto a 2D space. We plotted the UMAP for the four models and their respective classifier architecture pairs (Figure \ref{UMAP_output}). The separation of clusters of colored points suggests that the models were able to differentiate the antibody sequences from one another by assigning similar output tensors to similar antibody sequences. We certainly had a big smile, seeing that our models not only produced high-fidelity classifications but also showed some intelligent artistry.

\subsection{What do our models learn about the antibody sequence?}
We then wondered at what point were our models able to comprehend the biological implications hidden underneath the AA sequences. We hypothesized that AntiBERTa and BioBERT would likely do a better job than MSE2 and much better than GPT-2 at identifying biologically important antibody-specific features such as V gene usage, the somatic hypermutation counts, and antibody isotype.

\begin{figure}[ht]
\vskip 0.2in
\begin{center}
\centerline{\includegraphics[width=\columnwidth]{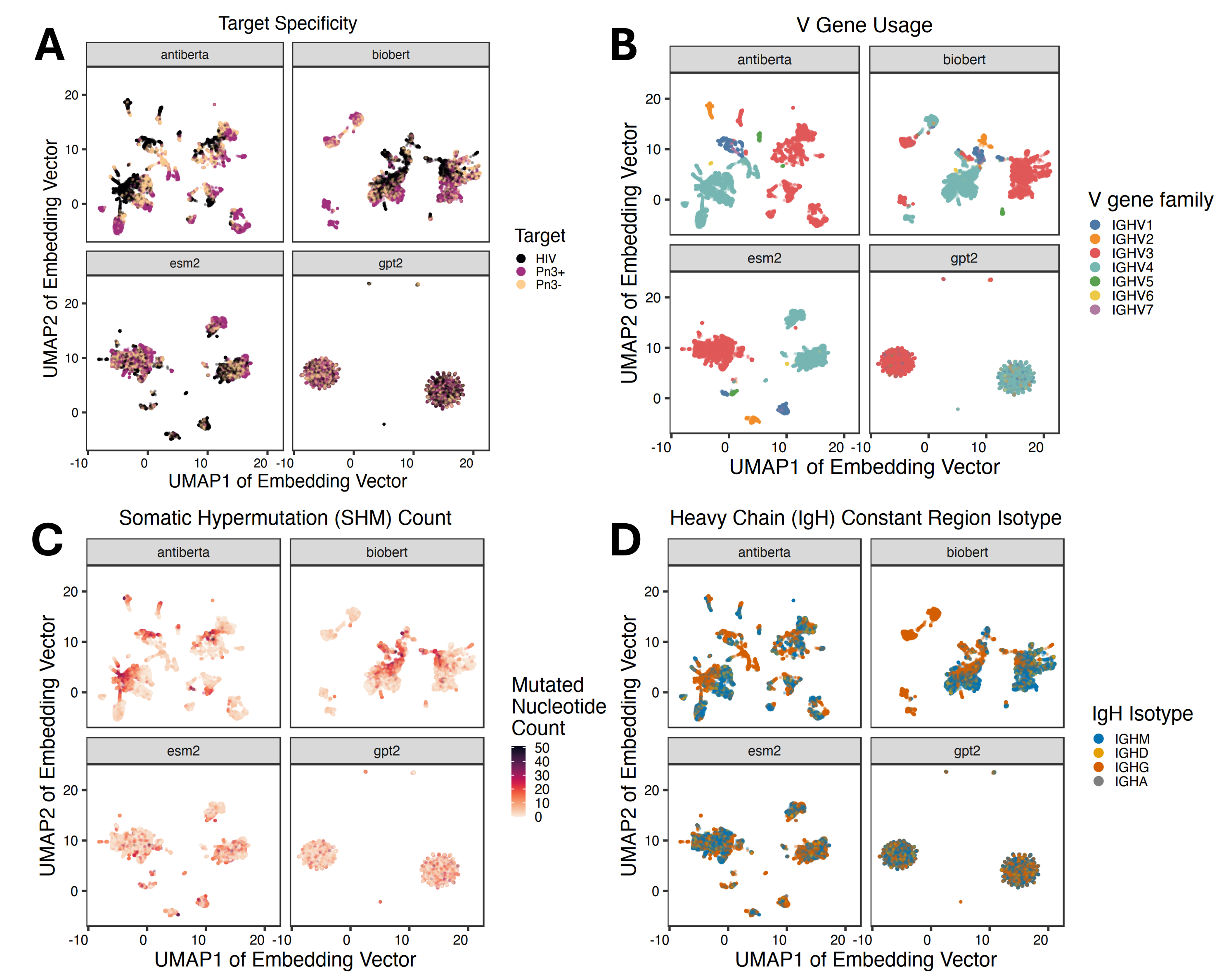}}
\caption{UMAP of embedding vectors under different language models highlighted with various antibody-specific biological properties. A. The antibody target specificity. B. The V gene family of each antibody sequence. C. The somatic hypermutation of each antibody sequence. D. The antibody heavy chain (IgH) constant region isotypes.}
\label{UMAP_embed}
\end{center}
\vskip -0.2in
\end{figure}

The UMAP projections of antibody sequence embeddings from AntiBERTa, BioBERT, ESM2, and GPT-2 reveal fascinating insights. AntiBERTa, BioBERT, and ESM2—having been specifically trained or tailored for protein and antibody-specific tasks—demonstrated a remarkable ability to capture biologically relevant features in their last hidden layers (Figure \ref{UMAP_embed}). Interestingly, AntiBERTa and BioBERT appeared to show differentiation of target specificity even at the embedding space before the following classifier model (Figure \ref{UMAP_embed}A).

In contrast, while capable of reflecting some sequence features like V gene usage, GPT-2—a model not specialized for antibodies—was not able to capture biologically meaningful distinctions amongst the sequences, showcasing that biological sequences carry semantics far different from natural languages. However, the classification outcomes were still decent, perhaps suggesting that antibody sequence comprehension was not necessary for our simple classification application. Nonetheless, this outcome reinforces our hypothesis: models specifically trained on protein or antibody data are better equipped to decode the complex interplay of sequence and function inherent to these biomolecules.

In the projections, AntiBERTa, BioBERT, and ESM2 embeddings beautifully delineate clusters of sequences based on V gene usage, a key determinant of antibody diversity Figure \ref{UMAP_embed}B). This suggests that these models internalize the sequence features associated with different gene families, effectively “understanding” this critical layer of antibody biology.

Further stratification based on somatic hypermutation (SHM) counts reveals another layer of complexity captured by these models (Figure \ref{UMAP_embed}C). As alluded to in the introduction, B cells mutate their BCR every time they divide, so the SHM counts reflect the immunological history of B cells and is the cornerstone metric for the dynamic improvement of antibody specificity and affinity—a process termed "affinity maturation". High SHM counts are typically associated with more effective antigen binding and neutralization, which is reflected by the visible correlation between high SHM sequences and HIV+ and Pn3+ sequences (Figure 5A \& 5C). By capturing and stratifying sequences based on SHM counts, AntiBERTa and BioBERT models demonstrate a nuanced grasp of not only the static sequence features but also the developmental process during antibody maturation.

The variable region of an antibody determines its target while the constant region determines the immunological outcome after binding to the target. The antibody heavy chain constant region has 5 isotypes (IgHM, IgHD, IgHG, IgHA, and IgHE), and each isotype will carry out distinct downstream effects. For example, IgM is often the first antibody produced during an immune response, while IgG is associated with long-term immunity and antigen neutralization. The sensitivity of AntiBERTa and BioBERT embeddings to IgH isotypes is particularly intriguing because the IgH sequences were not included at all in the training data (only the variable region sequences) (Figure \ref{UMAP_embed}D). The fact that AntiBERTa and BioBERT can infer isotype information—despite not having direct access to constant region sequences—suggests they have learned to pick up on subtle, biologically relevant patterns. We hypothesize that this sensitization may stem from the strong correlation between IgH isotypes and SHM counts (Figure \ref{UMAP_embed}C \& D). For instance, IgHG antibodies typically exhibit higher SHM rates compared to IgHM, because IgHM antibodies are first responders while IgHG antibodies have more time to affinity-mature.

This finding underscores an exciting aspect of deep learning in biological research: the ability to identify patterns and contexts that biologists or computer scientists might not immediately recognize. For example, while the link between SHM and isotype is well-documented, the ability of these models to deduce this relationship without direct isotype information points to their strength in uncovering latent, biologically significant structures.

\subsection{What do our models perceive as important motifs in an antibody sequence?}
\begin{figure}[ht]
\vskip 0.2in
\begin{center}
\centerline{\includegraphics[width=\columnwidth]{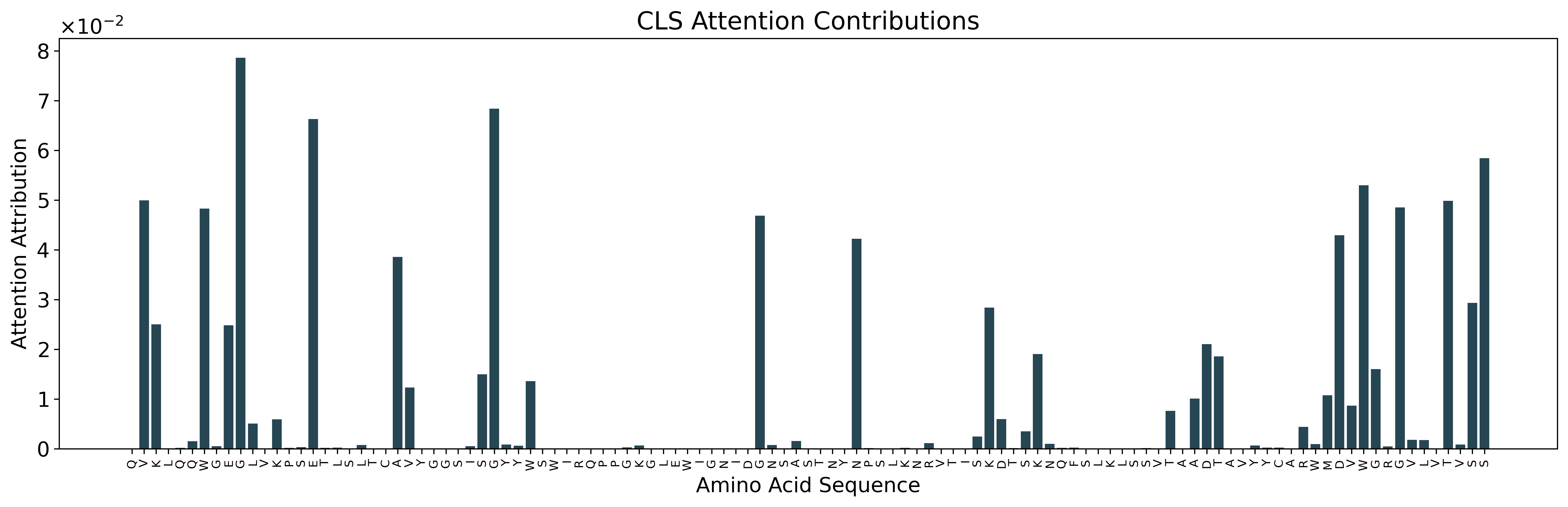}}
\caption{The attention contribution scores of the last hidden layer [CLS] token for each AA in a sample antibody sequence that ran through the ESM2-MLP model.}
\label{example_attribution_esm2_mlp}
\end{center}
\vskip -0.2in
\end{figure}

The next question that emerged was how the models were able to internalize the underlying biological context. To explore what the models consider important, we focused on extracting the attention attribution scores from the [CLS] token, a key feature in transformer models that aggregates the information across the entire sequence. We then asked: which amino acids at what positions attributes most to the [CLS] token? By examining these attention scores for every sequence in our dataset, we could identify regions that the models deem most relevant. To provide a primer for our analysis, we picked a sample antibody sequence that ran through the ESM2 with an MLP classifier, and we plotted the [CLS] token attention attribution scores from the last hidden layer (Figure \ref{example_attribution_esm2_mlp}). It seems like the 8 AAs with high scores in the first 30 AA sequence, the 3 AAs in the middle, and the 6 AAs in the last 15 AA sequence were important motifs most representative of this particular sequence.

\begin{figure}[ht]
\vskip 0.2in
\begin{center}
\centerline{\includegraphics[width=0.7\columnwidth]{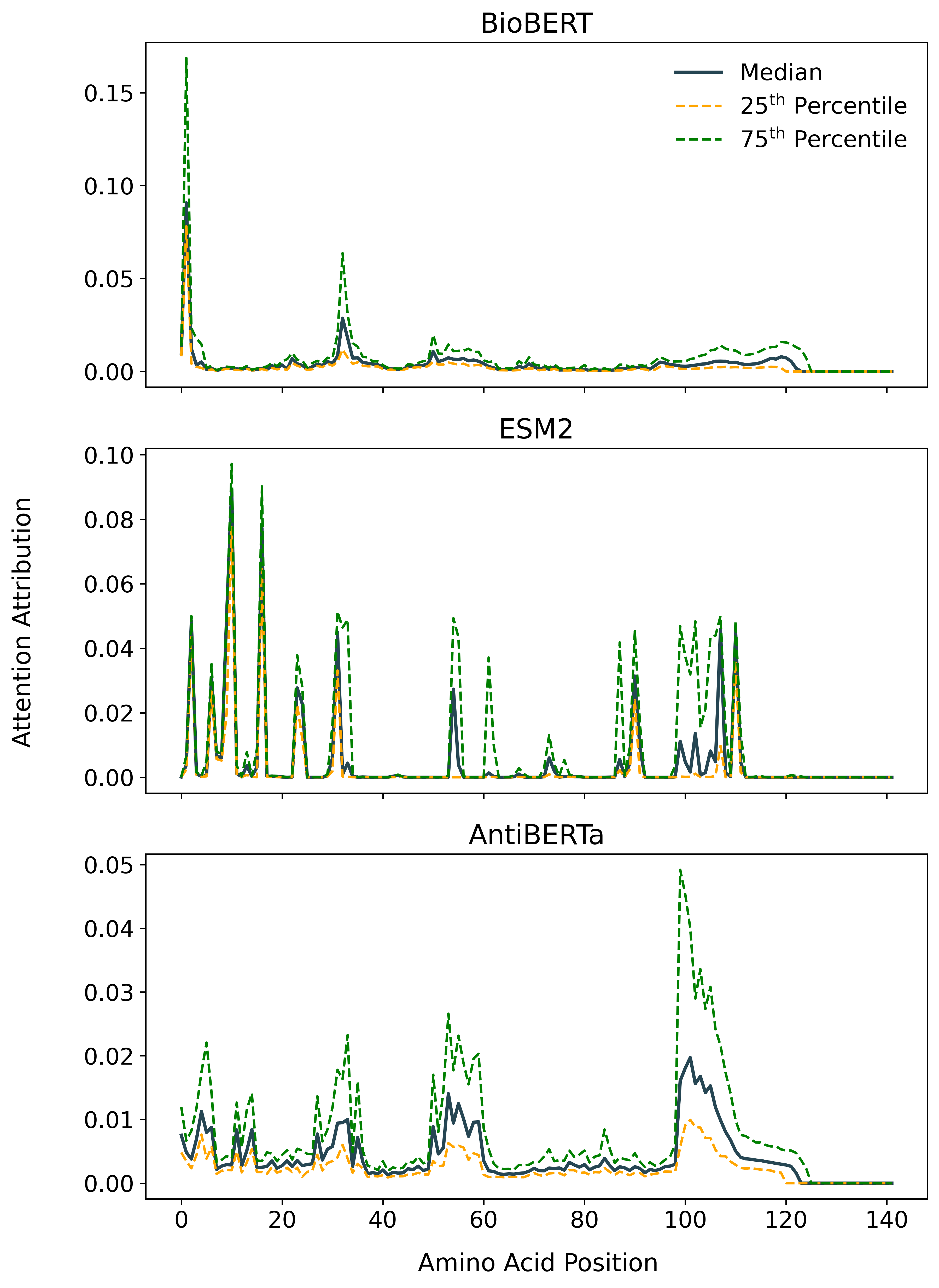}}
\caption{[CLS] token attention contribution scores for each AA of all sequences in the dataset. The median attention scores for each AA across all 10,377 sequences are plotted in the black line, and the 25th and 75th percentiles are plotted in dashed yellow and green lines, respectively.}
\label{all_attention}
\end{center}
\vskip -0.2in
\end{figure}

Next, we extrapolated the attention attribution scores of the last hidden layer [CLS] token from BioBERT-, ESM2-, and AntiBERTa-MLP models for all antibody sequences and plotted the median attention scores for each position along the antibody sequences (Figure \ref{all_attention}).

AntiBERTa, being fine-tuned specifically for antibody sequences, exhibited a clear pattern of attention to the first 15 AAs, a region most commonly used by bioinformaticians to declare V gene alignment. More notably, AntiBERTa's attention scores showed distinct peaks at the three CDRs. The attention score progressively increased from CDR1 to CDR3, which resonates with known immunology: CDR3, in particular, plays a dominant role in antigen binding. This suggests that AntiBERTa, with its specialized pre-training together with our antigen-specific classification training, understands the importance of these regions and could focus on them accordingly.

ESM2 and BioBERT, although not explicitly fine-tuned for antibodies, also demonstrated an ability to discern key antibody regions. They both assigned high attention scores to the early AAs in the sequence, which explains and corroborates their superb ability to differentiate sequences by their V gene usage (Figure \ref{UMAP_embed}B). Though not as prominent as AntiBERTa, the ESM2 model still showed a decent “understanding” that the CDRs are important regions with obvious attention score peaks.

From a biologist's perspective, these results are exciting. Antibodies rely on the precise configuration of their CDRs to recognize and bind antigens. The fact that these models were able to detect and assign attention to these regions confirms that the biological relevance of these segments is captured within the models' embeddings. For AntiBERTa, this aligns with our expectations given its antibody-specific fine-tuning, but the ability of ESM2 and BioBERT—trained on more general protein data—to still identify these key functional regions highlights the power of large-scale PLMs in understanding complex biological structures. By visualizing these embeddings and attention attributions, we catch a glimpse of how PLMs distill the language of proteins into meaningful, biologically relevant representations, bridging the gap between sequence data and antibody functionality.

\subsection{Would biological insights improve deep learning models performance?}
The ability of our models to identify biologically significant regions like the CDRs through training is certainly impressive and raises an interesting question: Could incorporating biological insights explicitly into the training process further enhance model performance? If models like AntiBERTa and ESM2 can infer the importance of CDRs independently, would focusing their training exclusively on these critical regions accelerate learning or improve downstream performance?

To test this, we decided to experiment with a biologically informed approach: rather than pooling embeddings across the entire sequence, we pooled only the embedding tensors corresponding to the CDR3 region for each sequence. Given that the CDR3 region is pivotal for antigen binding and rich in biological information, we hypothesized that this targeted pooling would minimally affect AntiBERTa, which already assigns significant importance to CDR3, but could have a more pronounced impact on ESM2 and BioBERT models, where attention to CDR3 is less pronounced. By aligning model training with known biological insights, we aim to explore whether this synergy between biology and deep learning can unlock additional performance gains.

When we compared validation accuracy as a function of the number of training samples, the results were striking (Figure \ref{pooling}). AntiBERTa, which already demonstrated a strong intrinsic focus on the CDR3 region, showed almost identical training accuracy curves between CDR3 pooling and full sequence pooling. This outcome aligns with our hypothesis: as a model fine-tuned for antibody sequences, AntiBERTa already prioritizes CDR3 as a critical region, and explicit focus on CDR3 during training provided no additional advantage.

\begin{figure}[ht]
\vskip 0.2in
\begin{center}
\centerline{\includegraphics[width=0.8\columnwidth]{images/pooling.png}}
\caption{The validation accuracy as a function of the number of training samples for BioBERT (blue), ESM2 (green), and AntiBERTa (salmon) models followed by FC classifier using CDR3 mean pooling (circle) or full-sequencing mean pooling (triangle).}
\label{pooling}
\end{center}
\vskip -0.2in
\end{figure}

In contrast, ESM2 exhibited a significant boost in training efficiency with CDR3 pooling. The validation accuracy plateau was reached far more quickly compared to full sequence pooling, suggesting that isolating the CDR3 region during training enabled the model to hone in on the most biologically relevant features without being diluted by the rest of the sequence. This result combined with observations from Figure 7 implies that while ESM2 can recognize the importance of CDR3 through the entire sequence, its training process benefits greatly from explicitly directing attention to this critical region.

BioBERT also showed improvement with CDR3 pooling, though the gains were less pronounced. We suspect that while BioBERT was capable of learning antibody features, it may not inherently assign as much weight to CDR3. By emphasizing CDR3 during training, BioBERT was able to learn more efficiently. However, the modest improvement reflects the model's comparatively lower sensitivity to antibody-specific patterns.

We hope our analysis and findings demonstrate the potential of crosslinking biological insights with the training of deep learning models. By aligning the training process with known biological principles, we can improve training efficiency and accelerate performance gains, especially for models that are less specialized or general-purpose. Furthermore, the observation that CDR3 pooling significantly enhanced ESM2 and BioBERT suggests that targeted training strategies can bridge the gap between general protein understanding and antibody-specific tasks. This synergy between biology and deep learning not only improves model performance but also provides a framework for designing more efficient, interpretable, and biologically informed computational tools.

\section{Conclusions}

Our exploration of PLMs and their ability to capture antibody-specific biological features highlights the remarkable progress at the intersection of deep learning and immunology. From identifying key antibody features like V gene usage and SHM counts to uncovering the critical importance of CDRs, models like AntiBERTa, BioBERT, and ESM2 demonstrate a sophisticated understanding of antibody biology.

By experimenting with biologically informed training strategies, such as CDR3 pooling, we uncovered compelling evidence that integrating prior biological knowledge can significantly enhance model performance and training efficiency. While specialized models like AntiBERTa showed little improvement due to their inherent sensitivity to CDR3, models like ESM2 and BioBERT benefited from this guided focus, particularly in terms of faster convergence and improved accuracy with fewer training samples.

Perhaps most excitingly, this work demonstrates that deep learning can not only emulate but also complement the biological understanding of antibodies. From capturing biologically relevant features without explicit guidance to leveraging curated insights to improve training, PLMs offer a dynamic toolkit for studying antibody sequences at unprecedented scale and depth. As we continue to refine these models, the collaboration between biology and deep learning holds tremendous promise for advancing immunology, vaccine development, and therapeutic antibody design.

\section*{Impact Statement}

This work has implications for both machine learning and therapeutic antibody development. From a machine learning perspective, our findings on architecture-induced biases could inform the design of more effective protein language models, potentially extending beyond antibody applications to other biological domains. In therapeutic development, improved understanding of how different architectures capture antibody properties could accelerate the discovery and optimization of therapeutic antibodies, potentially reducing development costs and time-to-market. However, we acknowledge that faster therapeutic development pipelines may have complex societal implications, including questions of equitable access to resulting treatments and the need for careful safety validation. We encourage future work to address these considerations while building upon our technical contributions.

\bibliographystyle{unsrt}  
\bibliography{myref}

\end{document}